%% file: PaperForReview.tex
\documentclass[10pt,twocolumn,letterpaper]{article}

\usepackage{cvpr}              

\usepackage{CJKutf8}

\usepackage{graphicx}
\usepackage{amsmath}
\usepackage{amssymb}
\usepackage{booktabs}
\usepackage{multicol}
\usepackage{multirow}
\usepackage{indentfirst}
\usepackage{float}
\usepackage{color}
\usepackage[dvipsnames]{xcolor}

\usepackage{tabularx}
\usepackage[pagebackref,breaklinks,colorlinks]{hyperref}

\usepackage[capitalize]{cleveref}
\crefname{section}{Sec.}{Secs.}
\Crefname{section}{Section}{Sections}
\Crefname{table}{Table}{Tables}
\crefname{table}{Tab.}{Tabs.}

\def\cvprPaperID{8799} 
\def\confName{CVPR}
\def\confYear{2023}
\usepackage{booktabs}

\begin{document}

\title{DoctorGLM: Fine-tuning your Chinese Doctor is not a Herculean Task}

\author{Honglin Xiong$^{1}$*\qquad 
Sheng Wang$^{1,2,3}$* \qquad Yitao Zhu$^{1}$* \qquad Zihao Zhao$^{1}$* \qquad Yuxiao Liu$^{1}$\\
Linlin Huang$^{4}$ \qquad Qian Wang$^{1,3}$ \qquad Dinggang Shen$^{1,3}$\\ 
$^1$ShanghaiTech University \qquad $^2$Shanghai Jiao Tong University   \\
$^3$United Imaging Intelligence \qquad $^4$Huashan Hospital, Fudan University \\
\{xionghl, wangqian2, dgshen\}@shanghaitech.edu.cn \qquad wsheng@sjtu.edu.cn
}

\maketitle


\begin{abstract}
The recent progress of large language models (LLMs), including ChatGPT and GPT-4, in comprehending and responding to human instructions has been remarkable. Nevertheless, these models typically perform better in English and have not been explicitly trained for the medical domain, resulting in suboptimal precision in diagnoses, drug recommendations, and other medical advice. Additionally, training and deploying a dialogue model is still believed to be impossible for hospitals, hindering the promotion of LLMs.  
To tackle these challenges, we have collected databases of medical dialogues in Chinese with ChatGPT's help and adopted several techniques to train an easy-deploy LLM. Remarkably, we were able to fine-tune the ChatGLM-6B on \textbf{a single A100 80G in 13 hours}, which means having a healthcare-purpose LLM can be very \textbf{affordable}.
DoctorGLM is currently an early-stage engineering attempt and contain various mistakes. We are sharing it with the broader community to invite feedback and suggestions to improve its healthcare-focused capabilities: \href{https://github.com/xionghonglin/DoctorGLM}{https://github.com/xionghonglin/DoctorGLM}.

\end{abstract}

\section{Introduction}


Large Language Models (LLMs) are highly advanced artificial intelligence systems that have undergone extensive training on vast amounts of text data. By utilizing deep learning techniques, these models are able to generate responses that resemble human-like speech, making them incredibly useful in a variety of tasks, such as language translation, question answering, and text generation. OpenAI's GPT series, among other LLMs, has exhibited remarkable results, and has the potential to revolutionize various industries, including marketing, education, and customer service. LLMs are highly sought after for their ability to process and understand large amounts of data, which makes them well-suited to solve complex problems.

Despite their remarkable performance in natural language processing, large language models like ChatGPT and GPT-4 have not been designed specifically for the medical domain. 
As a result, using LLMs for medical purposes may lead to suboptimal precision in diagnoses, drug recommendations, and other medical advice, potentially causing harm to patients.
Another limitation of large language models like ChatGPT and GPT-4 is that they are typically trained in English, which restricts their ability to comprehend and respond to other languages. This can create a barrier for individuals who do not speak English as their first language and limit the accessibility of medical advice to a wider audience. 
In order to overcome these limitations and better integrate LLMs into the lives of most ordinary people,  it's crucial to develop medical-tailored LLMs that can be trained in multiple languages. 
This will not only improve the accuracy of medical advice provided by these models but also make it more accessible to a wider audience.

In order to improve the precision and accuracy of medical advice provided by language models in the medical domain, a database of medical dialogues in Chinese has been compiled. This database contains information from a large number of patients, including their symptoms, recommended medications, and the necessary medical tests. The database has been created to provide language models with extensive medical knowledge and to enable them to generate more accurate and personalized responses to medical queries. By incorporating this knowledge, the hope is to improve the ability of language models to diagnose illnesses and provide better recommendations to patients, ultimately improving the quality of healthcare.

\begin{figure*}[!ht]
    \centering
    \includegraphics[width=0.98\textwidth]{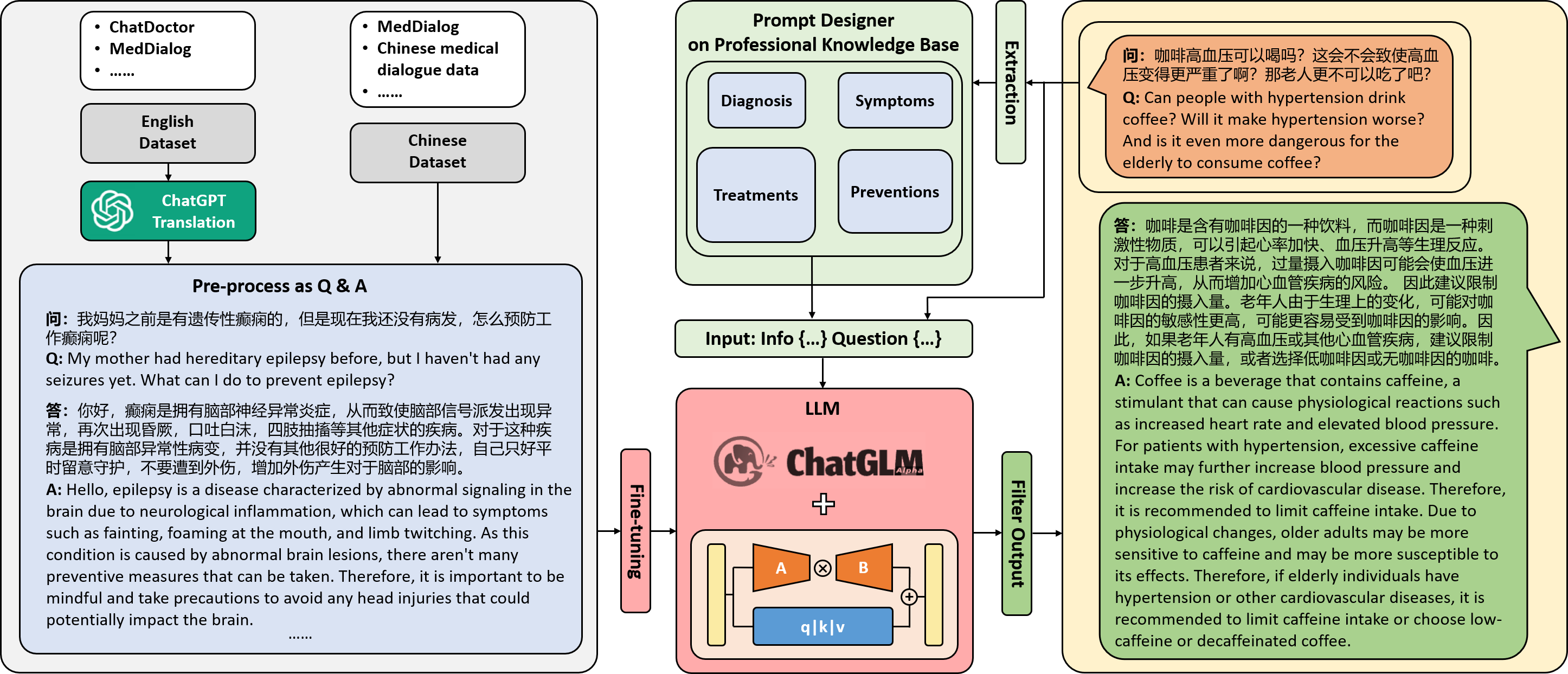}
    \caption{Overview of DoctorGLM fine-tuning and inference pipeline.}
    \label{usecases}
\end{figure*}

To optimize our medical language model for both Chinese and English languages and, more importantly, explore a feasible pipeline of customized medical LLMs, we fine-tuned it based on ChatGLM, a pre-trained language model with 6 billion parameters. This model is unique in that it is bilingual, offering proficiency in both English and Chinese. Furthermore, the GLM model has a unique scaling property that allows for INT4 quantization enabling effective inference on a single RTX 3060 (12G). This scaling property is a major breakthrough in the field of healthcare language modeling, as it allows for more efficient and cost-effective computation on affordable GPUs, making it easier for hospitals to deploy their medical dialogue models based on their in-house data.
Also, we use low-rank adaptation that facilitates fine-tuning on an A100 80G GPU. This allows for faster inference times, making it easier for researchers and developers to utilize large-scale language models for a variety of applications.

At present, the general public often assumes that large language models (LLMs) are monopolized by technology giants due to the substantial computational costs associated with ChatGPT. However, in this paper, we demonstrate that a specialized Chinese dialogue language model focused on the medical domain can be trained for less than 100 USD. We accomplish this by utilizing parameter-efficient tuning and quantization techniques, enabling the development of an LLM-based system that can be customized for specific tasks. The main contributions of this paper are summarized below:

\begin{itemize}
\item We present the first attempt at training a non-English healthcare LLM.

\item We develop a comprehensive pipeline for training dialogue models, applicable across different languages and adaptable to any specific clinical department. The source code is made available on GitHub.

\item We demonstrate that the costs of training and deploying a personalized LLM are affordable, thus encouraging hospitals to train their own LLMs based on in-house data with ease.
\end{itemize}

\begin{table*}[htbp]
  \centering
  \caption{Statistics of the used datasets. Chat means multi-round QA. Syn. means whether the data is generated by other language models. CMD. means Chinese medical dialogue.}
    \begin{tabular}{cccccccc}
    \toprule
    Dataset & Department & Language & Q\&A  & Chat  & Number & Syn.  & Size \\
    \midrule
    \multirow{5}[1]{*}{CMD.} & Surgical & \multirow{5}[1]{*}{CN} & \multirow{5}[1]{*}{\checkmark} & \multirow{5}[1]{*}{×} &116K       & \multirow{5}[1]{*}{×} & 52MB \\
          & Obstetrics and Gynecology &       &       &       & 229K  &       & 78MB \\
          & Pediatrics &       &       &       & 117K  &       & 47MB \\
          & Internal Medicine &       &       &       & 307K  &       & 102MB \\
          & Andriatria &       &       &       & 113K  &       & 44MB \\
    \midrule
    MedDialog & Multiple & CN\&EN & \checkmark     & \checkmark     & 3.4M  & ×     & 1.5GB \\
    ChatDoctor & Multiple & EN    & \checkmark     & ×     & 5.4K  & \checkmark     & 2.9MB \\
    HearlthcareMagic & Multiple & EN    & \checkmark     & ×     & 200K  & ×     & 216MB \\
    \bottomrule
    \end{tabular}%

  \label{tab:dataset}%
\end{table*}%

\section{Large Language Models in Healthcare}
\begin{figure}[!ht]
    \centering
    \includegraphics[width=0.45\textwidth]{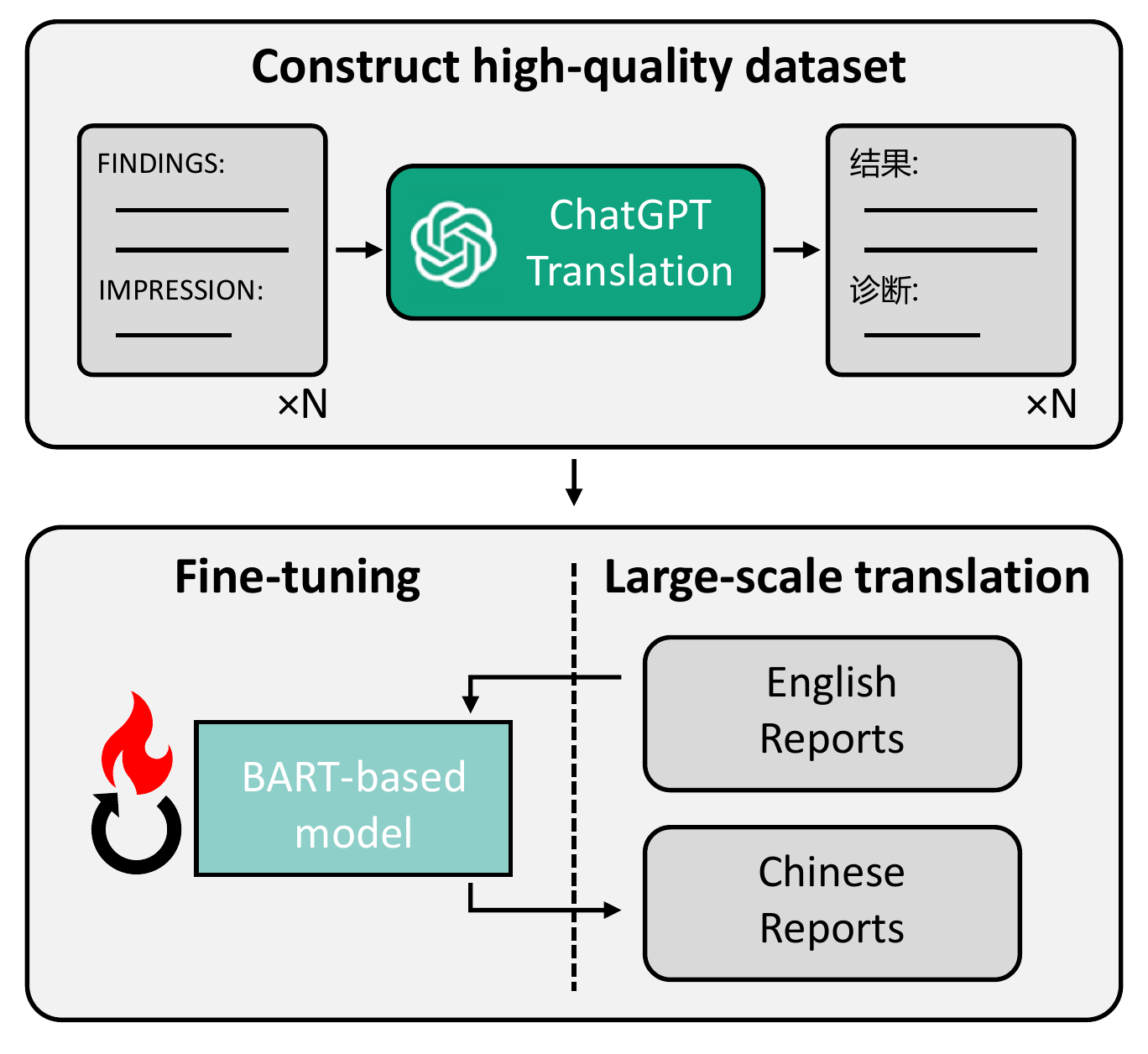}
    \caption{The implementation of large-scale translation. A tiny and high-quality dataset is built through ChatGPT. The collected dataset serves as a fine-tuning set for a pre-trained language model, enabling it to perform specialized machine translation.}
    \label{translation}
\end{figure}
Recent advances in Transformer architecture~\cite{vaswani2017attention} and computing power have enabled the training of large language models with billions of parameters, leading to a significant improvement in their ability to summarize, translate, predict and generate human-like text~\cite{brown2020language,singhal2022large,raffel2020exploring}. In pre-ChatGPT era, several healthcare language models have been developed based on general-purpose model weight and training schemes. BioBERT~\cite{lee2020biobert} and PubMedBERT~\cite{gu2021domain} are examples of BERT~\cite{devlin2018bert} models trained on PubMed for biomedical data, while ClinicalBERT~\cite{alsentzer2019publicly} was further trained on the MIMIC dataset and outperformed its predecessor. 

After ChatGPT showed the potential of 100B-scale model, researches expand healthcare language model to a much larger scale and give very promising results. Med-PaLM~\cite{singhal2022large} was developed in late 2022 using curated biomedical corpora and human feedback, and showed promising results, including a 67.6\% accuracy on the MedQA exam. ChatGPT, which was not given supplementary medical training, passed all three parts of the USMLE and achieved over 50\% accuracy across all exams and surpassed 60\% accuracy in the majority of them~\cite{kung2022performance}. ChatCAD~\cite{wang2023chatcad} combined medical image analysis models with ChatGPT and offered an interactive computer-aided diagnosis. ChatDoctor~\cite{yunxiang2023chatdoctor} is a medical chat model fine-tuned on LLaMA model using clinical QA that is synthesised by ChatGPT.

\section{Approach}

\subsection{Dataset with ChatGPT's Help}


It is worth noting that there exists a lot of high-quality datasets released in English. To utilize the available resources, we have translated  ChatDoctor~\cite{yunxiang2023chatdoctor} dataset 
to enhance the Chinese language proficiency of the DoctorGLM. 

The medical-targeted LLM requires professional training data, which asks high demands for English-Chinese translation. ChatGPT is capable of professional clinical text translation, but this would incur an overhead of tens of thousands of dollars for a large-scale dataset, which is unacceptable to most researchers. Here, we take a simple and low-cost approach to large-scale translation by leveraging the capabilities of ChatGPT.

Translation of the dataset is generally in two steps as shown in Figure~\ref{translation}. $X={\{x_1, x_2, ..., x_\textrm{N}\}}$ is initially selected from the ChatDoctor dataset, where $x_n$ is the raw English text, and corresponding high-quality translation $Y={\{y_1, y_2, ..., y_\textrm{N}\}}$ is obtained through ChatGPT API. Then, a BART-based pre-trained model~\cite{tiedemann-2020-tatoeba} is fine-tuned solely on paired $X$ and $Y$ without any additional datasets. In this way, the language model can distill the expert-level knowledge from ChatGPT and the refined small model can act as an acceptable alternative to LLMs~\footnote{Available at: \href{https://huggingface.co/zhaozh/medical_chat-en-zh}{https://huggingface.co/zhaozh/medical\_chat-en-zh}}.
We have translated ChatDoctor~\footnote{ChatDoctor: \href{https://drive.google.com/file/d/1nDTKZ3wZbZWTkFMBkxlamrzbNz0frugg/view?usp=sharing}{Google drive}} to use in our training.

To develop conversational models of high quality on a limited academic budget, ChatDoctor~\cite{yunxiang2023chatdoctor} utilized a strategy where each message from the disease database was entered as an individual prompt into the GPT3.5-turbo model to generate instruction data automatically. The prompts provided to the ChatGPT API contained the gold standard of diseases, symptoms, and drugs, resulting in a dataset that preserves the conversational fluency of ChatGPT while also achieving higher diagnostic accuracy than ChatGPT alone.


\subsection{Prompt Designer}
Large language models have achieved remarkable performance in conversational tasks. However, their outputs may be unreliable and deceptive. This issue also exists with ChatGLM, which is utilized in DoctorGLM. To address this problem and enhance the reliability of DoctorGLM's outputs, we use a prompt designer module that pre-processes the user's input.

The prompt designer module extracts relevant keywords such as the name of the disease or symptoms from the user's input. The module then utilizes the name of the most likely disease as a label and generates a brief description based on a professional disease knowledge library. This library comprises a comprehensive collection of detailed documents about various diseases. In particular, we have 3231 disease documents in detail, all of which are sourced from the Merck Manual of Diagnosis and Therapy.

The prompt designer's output includes information about the disease's symptoms, diagnosis, treatment options, and preventive measures. By providing a professionally generated prompt, the prompt designer expands the expertise and reliability of DoctorGLM for a particular disease. Additionally, it serves as a reliable source of information for users.

The generated prompt is integrated into the large language model, along with the original input, to improve the accuracy and reliability of DoctorGLM's responses. By using this approach, we can enhance the overall performance of DoctorGLM and provide reliable and trustworthy information to users. This system can also be used with other large language models that have decent in-context learning ability, e.g., ChatGPT, LLaMA and ChatGLM.


\subsection{Training of DoctorGLM}
We utilized the ChatGLM-6B model~\cite{du2022glm,zeng2023glm-130b} in developing our DoctorGLM. This open bilingual language model is based on the General Language Model (GLM) framework and has 6.2 billion parameters. ChatGLM-6B is optimized for Chinese QA and dialogue, and its technology is similar to ChatGPT. The model was trained on approximately 1 trillion tokens of Chinese and English corpus, with additional supervised fine-tuning, feedback bootstrap, and reinforcement learning using human feedback. Despite having only 6.2 billion parameters, ChatGLM-6B generates answers that are aligned with human preference. Furthermore, we use low-rank adaptation (LoRA) to finetune ChatGLM with only 7 million trainable parameters.

The fine-tuning process using all \textit{Chinese medical dialogue} dataset was conducted using an A100 GPU for a duration of 8 hours. The hyper-parameters employed in the training process were as follows: the batch size of 4, a learning rate of 2e-5 with lion optimizer, a total of 1 epochs, a maximum sequence length of 512 tokens, a maximum target length of 100 tokens. with no warmup and weight decay. The low-rank adaption is applied to $q,v$ and rank is set to 8 with alpha set to 16.

\section{Main Results}
\subsection{Compare to General Purpose Models}
Here we demonstrate some some QA with DoctorGLM in Table~\ref{tab:example1}, \ref{tab:example2} and \ref{tab:example34}. We compared to our base model ChatGLM-6B and ChatGPT (gpt-3.5-turbo).
DoctorGLM here have no additional prompt and filter. This model is trained based on CMD., with data from five departments. We also ask a doctor to mark the error and

In general, DoctorGLM tend to be too arbitrary while general models like ChatGPT are more conservative. For example, in Q2 (see Table~\ref{tab:example2}), ChatGLM suggested to use Ribavirin Granules,

\subsection{Generation Diversity}
Top-p and temperature are techniques used in text generation models to control the diversity and quality of the generated output.
Top-p (also known as nucleus sampling or softmax sampling) is a method used to select the most likely words based on their probabilities in the model's output distribution. It works by selecting the smallest set of words whose cumulative probability exceeds a certain threshold. This allows for more diverse output than simply selecting the most probable word at each step.
Temperature, on the other hand, is a scalar parameter that controls the randomness of the generated output. A higher temperature value results in more diverse and creative output, while a lower value leads to more conservative and predictable output. The temperature parameter controls the softmax function used to calculate the probabilities of the next word in the generated sequence. The results are presented on Table.~\ref{tab:ablation}

\subsection{How many iteration is good enough?}

For LoRA, we don't know yet.
See Table.~\ref{tab:iter} for more detail. During our training, we found model collapse after 50K iterations and can not produce any meaningful output.

\subsection{P-tuning vs. LoRA}
We also tested the P-tuning~\cite{liu2021p}, which is a method for fine-tuning large language models that involves optimizing only continuous prompts, significantly reducing storage and memory usage per task. It performs comparably to fine-tuning with only 0.1$\%$-3$\%$ of the fine-tuning parameters .
We compare LoRA and P-Tuning V2 for our model and the results are shown in Table.~\ref{tab:lora_and_ptuning}.


\input{examples}
\input{ablation}

\input{training_iter.tex}
\section{Cost}
Training a LLM from scratch with web-scale data can be a costly endeavor, but fine-tuning can be a more economical approach. DoctorGLM's training process can handle approximately 80,000 single question and answer pairs per hour per GPU. Assuming that three epochs are necessary, and the cloud computing server of an A100 GPU is approximately 5 USD per hour, the total training time required is 3.75 hours, which translates to a cost of approximately 18.75 USD for fine-tuning a DoctorGLM on 100,000 QA pairs.

On the other hand, the inference process for DoctorGLM requires only about 13 GB of GPU memory and can be performed on a consumer-level GPU such as an RTX 3090. This results in a total cost (inference PC) of approximately 1500 USD. 

It's worth noting that these costs may vary depending on the current price of GPUs and your location, but they remain relatively affordable for many research institution and hospitals.

\section{Technical Limitations}
\textbf{This work is in a very early stage and contains numerous mistakes, making it unsuitable for any commercial or clinical use.} One of the reasons we have published our work is to invite the broader community to help improve this healthcare-focused language model, with the aim of making it more accessible, affordable, and convenient for a larger audience. Below are some critical technical issues we encountered during this project:

\begin{enumerate}
\item DoctorGLM experiences a loss in capability during logistic training, and it occasionally repeats itself (see Table~\ref{tab:example1} A4 (DoctorGLM)). We suspect that fine-tuning typically incurs a higher alignment cost compared to reinforcement learning with human feedback (RLHF).
\item Generating a response takes approximately 15 to 50 seconds, depending on token length, which is significantly slower than interacting with ChatGPT via the web API. This delay is partly due to the chatbot's typing indicator.
\item We are currently facing difficulties in quantizing this model. While ChatGLM runs satisfactorily on INT-4 (using about 6G), the trained LoRA of DoctorGLM appears to have some issues. As a result, we are currently unable to deploy our model on more affordable GPUs, such as the RTX 3060 and RTX 2080.
\item We have noticed that the model's performance declines with prolonged training, but we currently lack a strategy for determining when to stop training. It appears that cross-entropy is an overly rigid constraint when fine-tuning LLMs.

\end{enumerate}


{\small
\bibliographystyle{ieee_fullname}
\bibliography{PaperForReview}
}
\newpage


\end{document}

%% file: examples.tex
\newcommand{\tred}[1]{\textcolor{red}{#1}}
\newcommand{\tgrn}[1]{\textcolor{ForestGreen}{#1}}
\newcommand{\tblu}[1]{\textcolor{MidnightBlue}{#1}}

\begin{CJK*}{UTF8}{gbsn}
\begin{table*}
  \begin{tabular}{|p{\textwidth}|}
    \hline
    \textbf{Q1} 近一个多月来，本人两肋、前胸包括胸口及两旁、后背两边、脊椎骨上段、包括扇子骨等处，就是上半身，发生不适隐痛，上半身很纠结，好像里面有手揪住一样，不舒展爽朗，十分难受，一会儿前胸和两肋不适难受，一会儿后背两边不适难受，同时发生隐痛。扭动身体，做做伸展活动，或者用拳头抵住揉搓不适隐痛处，身体就会舒展一些好受一些。总的来说就是上半身有肋骨的地方和上半段脊椎骨范围内发生不适和隐痛，一般同时两三个隐痛点，就是肋骨最下段的骨梢也发生不适隐痛。没有肋骨地方不痛，如肚子处就不隐痛。  本人于2011年11月做了胆囊摘除和胆总管结石手术。不是腹腔镜，是开了一条约18公分的大口子和两个导流小洞。此种不适难受隐痛情况2012年春天发生过一次，时间不长自动好了。但这次严重时间又长，不适和隐痛感觉有时呈块状，有时呈条状，一般同时有两三处不适隐痛，让人很难受，还有越来越重的感觉。 \\
    \hline
    \textbf{A1 (DoctorGLM)} 病情分析：您好，\tgrn{这种情况考虑是肋间神经炎引起的疼痛} \tblu{(还需与心绞痛、心脏神经官能症、胸膜炎、肋软骨炎等鉴别，可以根据鉴别诊断补充问诊，明确诊断)}
    
    指导意见：建议口服消炎药，止痛药，维生素B族，按疗程治疗，疼痛缓解后，练习抬胸运动，可缓解疼痛，除了正规治疗胆石症外，患者需要多咨询专家建议，和医生保持沟通，患者还需要重视护理方面，例如合理饮食，保持心情愉快与此同时患者还要注意选择一家专业医院诊治，这样才能得到良好的治疗效果 \\
    \hline
    \textbf{A1 (ChatGPT)} 很抱歉听到您最近出现了这些不适和隐痛的症状。根据您的描述，您的症状似乎涉及到上半身多个部位的肋骨和脊椎骨等骨骼结构，以及随之产生的不适和隐痛感觉。\tgrn{由于您之前做过胆囊摘除和胆总管结石手术，这些症状可能与手术本身有关，}\tblu{(距离手术时间很久了，后来自动缓解，中间未再发作，不太考虑与手术直接相关。)}但也有可能是其他原因引起的。因此，建议您及时就医，进行详细的身体检查和相关检查，以确定症状的具体原因和处理方法。
    您可以咨询家庭医生或内科医生，他们可以进行初步的检查和评估，并根据需要进行相应的检查，如X光、CT、MRI或血液检查等。在等待就医的过程中，您可以注意休息、保持良好的姿势、避免过度运动或重物提起，同时可以使用热敷或冷敷等方法缓解疼痛和不适感。
    最后，建议您保持乐观心态，积极配合医生治疗和康复。祝您早日康复！ \\
    \hline
    \textbf{A1 (ChatGLM)} 所述症状可能是由多种原因引起的，包括身体、心理和情绪等方面。根据所述症状，建议进行以下检查和测试：
    
    1. 医生体检：进行全面的身体检查，包括心肺听诊、血压测试、视力、听力、皮肤测试等，以确定任何潜在的健康问题。
    
    2. 心电图(EKG)：测试心脏健康，评估是否有潜在的心脏问题。
    
    3. 核磁共振(MRI)：测试骨骼和肌肉系统的健康，以及评估是否有潜在的神经问题。
    
    4. 血液检查：测试贫血、感染、肿瘤等问题。
    
    5. 心理健康测试：测试情绪和心理健康，例如压力测试、焦虑测试等。
    
    如果症状持续存在或变得更加严重，请尽快咨询医生以获取专业建议和治疗。\tblu{(过于笼统，没有针对性)}
    \\
    \hline

  \end{tabular}
  \caption{Doctor's comments are marked in \tblu{blue}. Factual errors are marked in \tred{red}. Improper diagnosis are marked in \tgrn{green}.}
  \label{tab:example1}
\end{table*}

\begin{table*}
  \begin{tabular}{|p{\textwidth}|}
    \hline
    \textbf{Q2} 医生我家宝宝4岁7个月.这几天连续五天发烧白天晚上都是.基本上都39度大约.消炎药也换过了.查血也查了.炎症又不太高结果医生说是急性扁桃体炎.很忧心求方法谢谢! \\
    \hline
    \textbf{A2 (DoctorGLM)} 孩子发热，一般考虑炎症的问题，建议给孩子多喝水，多喝水有助于降低体温，建议给孩子口服利巴韦林颗粒治疗，也可以口服小儿氨酚黄那敏颗粒治疗，可以口服消炎药，如阿莫西林克拉维酸钾等，建议平时注意孩子卫生，不要接触刺激物，多给孩子喝水，注意孩子体温变化，不要让孩子发烧，以上是对“宝宝得了急性扁桃体炎该怎么办？”这个问题的建议，期望对您有帮助，祝您健康！\tblu{(不这样建议的，先明确致病病原体。病毒：自限性，无需使用抗病毒药抗菌药，可以应用缓解症状的药物，利巴韦林一般用于呼吸道合胞病毒相关的肺炎，该药可能对儿童生长有影响。 细菌性则根据病原菌类型对应使用抗菌药，如A群β溶血性链球菌 选择阿莫西林、阿莫西林克拉维酸钾，支原体则选择阿奇霉素。还应特别关注相应的并发症。) } \\
    \hline
    \textbf{A2 (ChatGPT)} 很遗憾听到您家宝宝一直发烧，但是如果医生已经诊断出是急性扁桃体炎，那么您需要给宝宝提供适当的治疗和护理。
    以下是一些建议：
    
    1. 给宝宝多喝水，保持充足的水分摄入，可以帮助降低体温，并保持身体水分平衡。
    
    2. 让宝宝多休息，避免过度活动。可以适当减少学习和玩耍时间，帮助宝宝恢复体力。
    
    3. 您可以使用退烧药，如布洛芬或对乙酰氨基酚，以帮助控制宝宝的发烧症状。但一定要按照医生的建议使用。
    
    4. 给宝宝提供温暖和舒适的环境。确保房间内的温度适宜，并保持空气流通，以帮助宝宝呼吸更轻松。
    
    5. 给宝宝提供温和的饮食，以保持身体营养平衡。可以选择易于消化的食物，如米粥、面条和蒸蛋等。
    
    6. 如果宝宝的症状持续或加重，请及时与医生联系。可能需要进一步的检查或治疗。
    最后，希望您家宝宝早日康复，注意预防感染和保持健康的生活习惯。 \tblu{(过于笼统，没有帮助)}\\
    \hline
    \textbf{A2 (ChatGLM)} 急性扁桃体炎是儿童常见疾病之一，可能导致发热和喉咙疼痛等症状。根据你提供的信息，宝宝连续发烧五天，炎症不高，可能是急性扁桃体炎。
    以下是一些建议：
    
    1. 休息：宝宝在发烧时需要休息，避免过度活动和消耗体力。
    
    2. 饮食：给宝宝提供足够的水分和营养物质，以保证身体的恢复和修复。可以喂宝宝一些\tred{退烧汤}或粥，如鸡汤或蔬菜粥。
    
    3. 药物治疗：根据医生的建议使用消炎药，但是需要注意剂量和频率，以免产生不良反应。可以使用退烧药来降低宝宝的体温。
    
    4. 降温：可以使用冰袋或湿毛巾等方法来降温，但是需要注意不要将冰袋或湿毛巾放在宝宝的大脑或眼睛上，这可能会对造成伤害。
    
    5. 检查：最后，为了确定病因和诊断，需要再次给宝宝进行血液检查，以便确定是否感染病毒或其他病原体。
    
    要带宝宝去看医生，遵循医生的建议，进行必要的治疗和监测。同时，要积极配合医生的治疗，监测宝宝的病情变化，以便及时进行调整和治疗。
    
    \\\hline
    
    \textbf{Reference}  儿童急性扁桃体炎诊疗—临床实践指南（2016）
    
    1　一般治疗　卧床休息、清淡饮食、多饮水、加强营养及保持排便通畅；咽痛剧烈或高热时，可口服退热药及镇痛药。
    
    2　抗菌药物的使用　病毒性急性扁桃体炎常为自限性，无需使用抗菌药物治疗，可以考虑使用中药等治疗。
    A群β溶血性链球菌为本病的主要致病细菌，对于有细菌感染证据的急性扁桃体炎患儿， β内酰胺类为抗菌药物治疗的一线首选药物， 抗菌治疗应以清除病灶致病菌为目的， 疗程至少10 d， 根据病情轻重， 决定给药途径。青霉素类， 如： 阿莫西林、 阿莫西林+克拉维酸制剂等口服为推荐药物 。头孢类抗菌药物由于抗菌谱更广， 也可以作为一线药物治疗。对青霉素过敏的患儿或考虑为肺炎支原体感染者， 建议使用阿奇霉素等大环内酯类抗菌药物治疗， 阿奇霉素剂量为每日1次给药， 推荐使用剂量是10 mg/（kg·d）， 连续使用 3 d为1个疗程； 也有使用 5 d疗程的用法： 首剂10 mg/（kg·d）， 第2～5天5 mg/（kg·d）。或者12 mg/（kg·d），连续使用5 d为1个疗程。
    \\
    \hline

  \end{tabular}
  \caption{Doctor's comments are marked in \tblu{blue}. Factual errors are marked in \tred{red}. Improper diagnosis are marked in \tgrn{green}.}
  \label{tab:example2}
\end{table*}

\begin{table*}
  \begin{tabular}{|p{\textwidth}|}
    \hline
    \textbf{Q3} 每天清晨第一次尿完之后就一直感觉到尿频。。去尿就一点点。。到中午就回好了。但是尿尿的时候感觉到尿急。。憋不住的那种。这是怎么回事？ \\
    \hline
    \textbf{A3 (DoctorGLM)} 您好，考虑是\tgrn{前列腺炎}\tblu{(不这么直接考虑，可以提供多种考虑，建议进一步排查)}，前列腺炎多表现为排尿不适，尿频、排尿时尿道灼热、疼痛并伴随排尿困难，大便时尿道口有白色分泌物，甚至射精痛，阴茎痛，性欲减退，性交不适，便秘或排便痛，和他性勃起和射精时疼痛加重，建议最好去医院详细检查一下，明确确诊后再对症治疗，祝您健康
    \\
    \hline
    \textbf{A3 (ChatGPT)} 这种情况可能是尿路感染的症状之一，也可能是其他泌尿系统疾病引起的。建议您去看医生进行检查。

    尿路感染是由细菌感染尿道、膀胱或肾脏等泌尿系统器官引起的感染。尿频、尿急、尿痛是尿路感染的常见症状之一，其他还可能包括尿色深、尿中带血、腰痛等。
    
    除了尿路感染外，其他可能引起类似症状的疾病还包括前列腺炎、膀胱炎、尿道狭窄、尿道结石等。这些疾病也需要进行相应的治疗，因此建议及时就医，以确定病因并得到正确的治疗。 \\
    \hline
    \textbf{A3 (ChatGLM)} 引起尿频和尿急的原因有很多，包括：
    
    1. 膀胱感染或细菌感染：膀胱感染或细菌感染可能会导致尿频、尿急和尿痛等症状。建议进行尿液检查以确定感染的类型。
    
    2. 膀胱过度活动\tblu{(膀胱过度活动症)}：膀胱过度活动是一种常见的症状，会导致尿频和尿急。如果膀胱过度活动受到关注，医生可能会建议使用一些药物来减轻症状。
    
    3. 前列腺问题：前列腺问题也可能导致尿频、尿急和尿痛等症状。如前列腺增生或前列腺癌，它们可能会影响尿道和膀胱。建议进行超声波检查或医生进行详细的诊断。
    
    4. 饮食和饮水：饮食和饮水量也可能对尿频和尿急产生影响。过量饮酒或咖啡因摄入、高糖饮食、饮食不当等都可能导致尿频。建议减少这些食品的摄入。
    
    如果症状持续或加重，建议咨询医生进行详细检查以确定病因。
    \\
    \hline
    \hline
    \textbf{Q4} 我有蚕豆病溶血的，现在我有一些症状会表现出出的的.您好蚕豆病溶血有什么症状？\\
    \hline
    \textbf{A4 (DoctorGLM)} 你好，\tred{蚕豆病是溶血性贫血的简称}\tblu{(该描述有误，蚕豆病是葡萄糖-6-磷酸脱氢酶（G6PD）缺乏症的一个类型。溶血性贫血有很多种)}，是由溶血性贫血、溶血和蚕豆病等疾病组成的一组疾病，蚕豆病的症状有：发热、腹痛、腹泻、恶心、呕吐、头痛、恶心、干呕、发热、发热、呕吐、腹泻、恶心、干呕等，此外，蚕豆病患者在救治期间，患者的饮食状况和护理也是很重要的，患者一定要留意歇息，可以适当活动，以免引发不必要的伤害，期望上述的答案可以帮助到你，谢谢 \\
    \hline
    \textbf{A4 (ChatGPT)} 蚕豆病溶血是一种罕见的遗传性疾病，主要由于摄入蚕豆或类似食物导致红细胞溶解而引起。以下是一些可能表现出的症状：

    1. 贫血：由于红细胞溶解，可能会导致贫血，这可能会导致疲劳、头晕、心悸等症状。
    
    2. 黄疸：红细胞溶解后，可能会释放出大量的胆红素，导致皮肤、黏膜变黄，眼睛也会发黄。
    
    3.腹痛、腹泻：在严重的情况下，溶血可能会导致胆汁淤积，引起腹痛和腹泻。
    
    4.尿中出现血液：红细胞溶解后，血红蛋白可能会在肾脏中过滤，导致尿液呈现出棕红色。
    
    5. 肝脾肿大：长期患病可能会导致肝脾肿大，引起腹胀、腹部不适等症状。
    
    如果您怀疑自己有蚕豆病溶血，请立即咨询医生。治疗通常包括避免摄入引起疾病的食物，输注红细胞等支持治疗。 \\
    \hline
    \textbf{A4 (ChatGLM)} 蚕豆病是一种由溶血性贫血引起的疾病，\tred{其病因是由于体内缺乏一种叫做“抗-G5520”抗体的物质}\tblu{(有误，G6PD缺乏。G6PD缺乏症发病原因是由于G6PD基因突变，导致该酶活性降低，红细胞不能抵抗氧化损伤而遭受破坏，引起溶血性贫血。)}。这种物质是蚕豆(一种豆类)引起的，如果身体无法正常产生这种抗体，就会导致溶血性贫血。
    
    蚕豆病的症状包括：
    
    1. 溶血性贫血：由于体内的红细胞被破坏，导致贫血，症状包括疲劳、虚弱、头晕、头痛、血红蛋白尿等。
    
    \tred{2. 皮肤反应：许多人会出现皮肤的变化，例如皮肤变黑或变红，眼睛周围的皮肤也可能变成紫色或黑色。}
    
    3. 胃肠道反应：许多人会出现恶心、呕吐、腹泻等胃肠道症状。
    
    4. 心脏问题：有时溶血性贫血会导致心脏问题，例如心脏肌肉受损或心脏血管问题。
    
    如果觉得自己出现了与溶血性贫血相关的症状，建议立即咨询医生，因为溶血性贫血可以非常危险，可能导致死亡。
    \\
    \hline
    
  \end{tabular}
  \caption{Doctor's comments are marked in \tblu{blue}. Factual errors are marked in \tred{red}. Improper diagnosis are marked in \tgrn{green}.}
  \label{tab:example34}
\end{table*}

\begin{table*}
  \begin{tabular}{|p{\textwidth}|}
    \hline
    \textbf{Q1} 近一个多月来，本人两肋、前胸包括胸口及两旁、后背两边、脊椎骨上段、包括扇子骨等处，就是上半身，发
生不适隐痛，上半身很纠结，好像里面有手揪住一样，不舒展爽朗，十分难受，一会儿前胸和两肋不适难受，一
会儿后背两边不适难受，同时发生隐痛。扭动身体，做做伸展活动，或者用拳头抵住揉搓不适隐痛处，身体就会
舒展一些好受一些。总的来说就是上半身有肋骨的地方和上半段脊椎骨范围内发生不适和隐痛，一般同时两三个
隐痛点，就是肋骨最下段的骨梢也发生不适隐痛。没有肋骨地方不痛，如肚子处就不隐痛。本人于2011年11月做
了胆囊摘除和胆总管结石手术。不是腹腔镜，是开了一条约18公分的大口子和两个导流小洞。此种不适难受隐痛
情况2012年春天发生过一次，时间不长自动好了。但这次严重时间又长，不适和隐痛感觉有时呈块状，有时呈条
状，一般同时有两三处不适隐痛，让人很难受，还有越来越重的感觉。 \\
    \hline
    \textbf{A1 (lora)} 您好，这种情况考虑是肋间神经炎引起的疼痛 (还需与心绞痛、心脏神经官能症、胸
膜炎、肋软骨炎等鉴别，可以根据鉴别诊断补充问诊，明确诊断)
指导意见：建议口服消炎药，止痛药，维生素B族，按疗程治疗，疼痛缓解后，练习抬胸运动，可缓解疼痛，除了
正规治疗胆石症外，患者需要多咨询专家建议，和医生保持沟通，患者还需要重视护理方面，例如合理饮食，保
持心情愉快与此同时患者还要注意选择一家专业医院诊治，这样才能得到良好的治疗效果
    \\
    \hline
    \textbf{A1 (ptuning)} 你好，你的情况考虑是肋间神经痛或肋软骨发炎。肋间神经痛是指由于长期的精神、体力活动等因素，引起肋间神经的经常性疼痛，并呈持续性或阵发性发作。肋软骨发炎是指由于局部感染，引起肋软骨发炎，表现为局部疼痛，伴有红肿热痛，疼痛明显，有脓性分泌物。 \\
    \hline
    \hline
    \textbf{Q2} 医生我家宝宝4岁7个月.这几天连续五天发烧白天晚上都是.基本上都39度大约.消炎药也换过了.查血也查了.炎
症又不太高结果医生说是急性扁桃体炎.很忧心求方法谢谢! \\
    \hline
    \textbf{A2 (lora)} 孩子发热，一般考虑炎症的问题，建议给孩子多喝水，多喝水有助于降低体温，建议给孩子口
服利巴韦林颗粒治疗，也可以口服小儿氨酚黄那敏颗粒治疗，可以口服消炎药，如阿莫西林克拉维酸钾等，建议
平时注意孩子卫生，不要接触刺激物，多给孩子喝水，注意孩子体温变化，不要让孩子发烧，以上是对“宝宝得了
急性扁桃体炎该怎么办？”这个问题的建议，期望对您有帮助，祝您健康！
    \\
    \hline
    \textbf{A2 (ptuning)} 急性扁桃体炎的主要症状是咽痛.发热.全身不适.咽喉部有弥漫性红肿.扁桃体实质呈充血.水肿.发炎.可伴发淋巴结肿大.扁桃体有脓性分泌物.症状在清晨起床前最明显.如果的药物效果不理想的话.建议可以手术治疗的.平时注意饮食清淡，多吃水果蔬菜。 \\
    \hline
    \hline
    \textbf{Q3} 每天清晨第一次尿完之后就一直感觉到尿频。。去尿就一点点。。到中午就回好了。但是尿尿的时候感觉到尿
急。。憋不住的那种。这是怎么回事？ \\
    \hline
    \textbf{A3 (lora)} 您好，考虑是前列腺炎，前列腺炎多表
现为排尿不适，尿频、排尿时尿道灼热、疼痛并伴随排尿困难，大便时尿道口有白色分泌物，甚至射精痛，阴茎
痛，性欲减退，性交不适，便秘或排便痛，和他性勃起和射精时疼痛加重，建议最好去医院详细检查一下，明确
确诊后再对症治疗，祝您健康
    \\
    \hline
    \textbf{A3 (ptuning)} 您好，这种情况主要考虑是泌尿系感染、前列腺炎、前列腺增生、前列腺炎等，建议到医院泌尿科做相关检查，根据检查结果采取相应的治疗。平时注意多喝水，不要吃辛辣食物。 \\
    \hline
    \hline
    \textbf{Q4} 我有蚕豆病溶血的，现在我有一些症状会表现出出的的.您好蚕豆病溶血有什么症状？ \\
    \hline
    \textbf{A4 (lora)} 你好，蚕豆病是溶血性贫血的简称，是由溶血性贫血、溶血和蚕豆病等疾病组成的一组疾病，蚕豆病的症状有：
发热、腹痛、腹泻、恶心、呕吐、头痛、恶心、干呕、发热、发热、呕吐、腹泻、恶心、干呕等，此外，蚕豆病
患者在救治期间，患者的饮食状况和护理也是很重要的，患者一定要留意歇息，可以适当活动，以免引发不必要
的伤害，期望上述的答案可以帮助到你，谢谢
    \\
    \hline
    \textbf{A4 (ptuning)} 蚕豆病溶血，是体内抗溶血抗体的自动免疫反应，在蚕豆病溶血患者体内可再次出现溶血反应，导致红细胞破坏，从而再次出现贫血和出血等症状，严重的可再次出现休克和心力衰竭，患者应当立即就诊，必要时采用抗溶血药物，必要时可行红细胞再生计数，血液净化，血液抗凝等救治 \\
    \hline

  \end{tabular}
  \caption{Comparison of two finetuning method: lora and ptuningv2.}
  \label{tab:lora_and_ptuning}
\end{table*}

\end{CJK*}

%% file: ablation.tex

\begin{CJK*}{UTF8}{gbsn}
\begin{table*}
  \begin{tabular}{|p{0.03\textwidth}|p{0.3\textwidth}|p{0.3\textwidth}|p{0.3\textwidth}|}
    \hline
    
     & Top-p = 0.6 & Top-p = 0.7 & Top-p = 0.8\\    
     \hline
    T = 0.90 & 孩子发烧的情况可能是感冒，细菌感染，病毒感染，气候变化等原因引起的可以给孩子吃点温白开水，多喝点温水，可以吃点抗病毒口服液，四季抗病毒口服液，阿奇霉素分散片等，如果体温超过38.5°C，可以吃布洛芬退热，要是体温低于38.5°C的话，可以吃退热药如布洛芬退热，多喝水，吃点温热的食物，忌辛辣刺激食物，注意休息，注意卫生，祝孩子早日康复 \tred{(F)}
    & 你好，你的情况是急性扁桃体炎，建议应用消炎药物治疗，如阿莫西林等，同时注意多喝水，饮食清淡，避免辛辣刺激食物另外可以应用退热药物如布洛芬等，同时可以应用清热解毒的中药治疗，如清热解毒口服液，清开灵等，如果发烧超过38.5度，可以应用退烧药如布洛芬等，超过39度，可以使用退烧药，如布洛芬等，超过40度，需要手术治疗 \tred{(C)}
    & 孩子急性扁桃体炎是儿童最常见的传染病之一其病因可能是细菌或病毒感染多由扁桃体炎病毒或细菌感染，或细菌感染后病毒复制通常有发热，头痛，不适，咳嗽，喉咙疼痛，喉咙有脓样或粘稠的分泌物可以口服阿莫西林克拉维酸钾，多饮温水，多休息必要时输液治疗 \tred{(B)}\\
    \hline
    T = 0.95 & 急性扁桃体炎是儿童常见疾病，由细菌感染引起，临床诊断上多表现为发烫，头痛，全身不适，咳嗽，咳嗽后咳加重，发烧后发烫，发烧后咳嗽，咳嗽后咳加重，有痰，嗓子红，喉咙痛，有咽痛，头痛，全身不适，发烧，咳嗽，咳嗽后咳加重，有痰，嗓子红，喉咙痛，有咽痛，发烧，发烧后发烫，发烧后咳嗽，咳嗽后咳加重，有痰，嗓子红，喉咙痛，有咽痛，发烧，发烧后发烫，发烧后咳嗽，咳嗽后咳嗽加重，有痰，嗓子红，喉咙痛，有咽痛，发烧，发烧后发烫，发烧后咳嗽，咳嗽后咳嗽加重，有痰，嗓子红，喉咙痛，有咽痛，发烧，发烧后发烫，发烧后咳嗽，咳嗽后咳嗽加重 \tred{(F)}
    & 急性扁桃体炎是小儿最常见的传染病之一，儿童及青少年的多发期，常伴发其他传染病，也可单发，由细菌或病毒感染引来，以发热及咽痛为主要症状，可伴发全身不适，咳嗽，头痛，乏力等治疗建议：1、病因治疗：根据病因不同，选择不同的治疗方案，如急性扁桃腺感染需给予抗生素治疗2、对症治疗：发热及咽痛者，可给与退热药如布洛芬等3、局部治疗：急性扁桃腺感染可给与局部治疗，如局部清洗，涂药等4、全身治疗：如急性扁桃腺感染严重，可用抗生素治疗，如阿莫西林克拉维酸钾，如青霉素类，红霉素等5、护理：适当休息，多饮水，避免刺激性饮食如扁桃体局部疼痛明显，可给与局部冷敷，如咽痛严重，可给与局部麻醉药，如利咽灵，速迪平等 \tred{(A)}
    & 你好，孩子发烧的原因有许多，比如肺炎、细菌或病毒感染，孩子由于免疫系统尚未完全发育，所以容易再次发生感冒，除了一些特殊疾病，比如急性肾炎、急性肺炎、急性支气管炎等，都有可能引发孩子发烧所以，最好到正规大医院儿科进行详细检查，诊断明确后，再实施治疗，以免延误病情注意给孩子防寒保暖，多让孩子喝水 \tred{(A)}\\
    \hline
    T = 0.99 & 你好，根据你的叙述，考虑是扁桃体炎，可以吃点阿莫西林，小儿氨酚黄那敏颗粒，多喝点水，注意卫生，慢慢会好的指导意见：建议你平时注意宝宝卫生，勤换内衣，多喝水，注意体温变化，必要时输液治疗 \tred{(A)}
    & 您好，您孩子的情况是急性扁桃体炎，急性扁桃体炎是儿童急性感染性疾病中最常见的病原体感染，主要是细菌感染如果症状不缓解，应尽早去医院就诊一般抗生素治疗1周左右，如病情不缓解，应改用抗病毒药物如孩子有高热，可先用退热药如果高热持续，可加用退热药此外，儿童扁桃体炎的确诊需要检查血液，检查孩子是否感染了病毒，细菌等病原体孩子高热时，家长应给孩子采取降温措施，避免孩子高热引发其他并发症 \tred{(B)}
    & 您好，急性扁桃体炎，一般建议先对症治疗，然后根据情况决定抗炎药物治疗，如果发烧超过38度，则可以使用退烧药，发烧后可以考虑用冰袋冷敷，防止发烧再次上升，注意多喝白开水，必要时可以喝一些抗病毒的口服液如果炎症不消除，就需要输液治疗，药物建议看医生祝你健康 \tred{(C)}\\
    \hline

  \end{tabular}
  \caption{The impact of different hyperparameter settings on the generated results, where T refers to temperature. Temperature and top-p are used to control the diversity and fluency of the generated text. We ask doctor to give a \tred{score (ABCF, from good to bad)}.}
  \label{tab:ablation}
\end{table*}
\end{CJK*}

%% file: training_iter.tex
\begin{CJK*}{UTF8}{gbsn}
\begin{table*}
  \begin{tabular}{|p{\textwidth}|}
    \hline
    \textbf{Q2} \& \textbf{Q$_{general}$}:中国的首都是哪座城市
    \\
    \hline
    \textbf{iter .5K} \textbf{A2}: 您好，扁桃体反复发炎的情况建议可以去医院行扁桃体切除手术，一般术后一周左右就可以恢复，术后可以口服抗生素一周左右，也可以去医院行口服抗生素一周的复查，如果恢复良好，可以慢慢减量口服抗生素，也可以行口服抗生素一个月左右的复查，如果恢复良好，可以逐渐停药。

    \textbf{A$_{general}$}: 中国的首都是北京。
    \\
    \hline
    \textbf{iter 5K} 
    \textbf{A2}: 你好，根据你的描述，宝宝这是急性炎症引起的症状，需要积极的进行治疗的，可以口服消炎药物和外用消炎的药物来进行治疗，平时需要多喝水，多吃蔬菜水果，加强营养，注意保暖，避免寒冷，必要时可以服用抗生素来进行治疗。，对于宝宝发烧的问题，建议家长朋友们可以及时对症治疗，不要让宝宝长时间发烧，否则的话会给宝宝带来不必要的伤害，必要的时候是需要及时就诊，希望家长们可以根据宝宝的情况选择对症治疗，这样有利于宝宝康复。

    \textbf{A$_{general}$}: 在北京、上海、南京、深圳、成都、重庆、广州、深圳、厦门、杭州、成都、青岛、济南、8、8、8、8、8、8、8、8、8、8、8、8、8、8、8
    \\
    \hline
    \textbf{iter 10K} 
    \textbf{A2}:病情分析：根据你描述的症状，考虑是急性扁桃体炎引起的发热，建议积极消炎治疗，多喝水，多休息，注意个人卫生，饮食清淡，避免辛辣刺激食物，多喝水，注意保暖，避免生冷油腻的食物，祝早日康复。指导意见：考虑是急性扁桃体引起的发热，建议尽早就医，在医生指导下治疗，平时注意饮食，不要吃辛辣刺激性食物，多喝水，多吃蔬菜水果，多喝水，避免劳累，避免感冒，多吃高蛋白，高热量的食物，补充能量，增强抵抗力，避免传染，祝你早日康复，身体健康。

    \textbf{A$_{general}$}: 中国的首都是北京市。
    \\
    \hline
    \textbf{iter 20K} 
    \textbf{A2}: 你好，根据你叙述的情况，你的情况是扁桃体炎引起的，你的情况是急性的，可以应用一些消炎药的，另外也可以应用一些清热解毒的药物治疗，注意保持好个人的卫生，多饮水，多吃水果和蔬菜，祝早日康复
    
    \textbf{A$_{general}$}: 中国的首都是北京市，北京市的简称是北京市，简称北京，是中国的首都。
    \\
    \hline
    \textbf{iter 30K} 
    \textbf{A2}: 你好，根据你的症状，考虑是扁桃体炎。扁桃体炎是细菌感染引起的常见疾病，是细菌感染引起的，主要是用抗生素治疗的，效果比较理想的，可以服用阿莫西林胶囊、利咽颗粒剂、咽红冲剂等治疗，同时配合扁桃体抽液、扁桃体内脓肿、淋巴结肿大等治疗。
    
    \textbf{A$_{general}$}: 中国的首都是北京市。
    \\
    \hline
    \textbf{iter 40K} 
    \textbf{A2}: 你好，根据你描述的情况可能是患有扁桃体炎的。扁桃体炎是细菌感染引起的，可以服用阿莫西林胶囊、阿奇霉素、红霉素等药物进行治疗。
    
    \textbf{A$_{general}$}: 你好，北京是首都。
    \\
    \hline
    \textbf{iter 50K} \textbf{A2}:治疗。
    
    \textbf{A$_{general}$}: 。。、
    \\
    \hline

  \end{tabular}
  \caption{LoRA, Generated response with different training iteration number.}
  \label{tab:iter}
\end{table*}

\end{CJK*}